%% file: main.tex
\documentclass{svproc}
\usepackage{authblk}
\usepackage{graphicx} 

\usepackage{hyperref}
\usepackage{url}

\begin{document}

\mainmatter 

\title{Team Delft's Robot Winner of the Amazon Picking Challenge 2016}

\author{Carlos Hernandez \inst{1} \and Mukunda Bharatheesha\inst{1} \and
Wilson Ko\inst{2} \and Hans Gaiser\inst{2} \and
Jethro Tan\inst{1} \and Kanter van Deurzen\inst{2} \and Maarten de Vries\inst{2}
\and Bas Van Mil\inst{2} \and Jeff van Egmond\inst{1}
\and Ruben Burger\inst{1} \and Mihai Morariu\inst{2} \and Jihong Ju\inst{1} \and
Xander Gerrmann\inst{1} \and Ronald Ensing\inst{2} \and Jan Van
Frankenhuyzen\inst{1} \and Martijn Wisse\inst{1,2}}

\authorrunning{Carlos Hernandez et al.} 

\institute{Robotics Institute, Delft University of Technology,\\
Mekelweg 2 2628 CD Delft, The Netherlands\\
\and
Delft Robotics, B.V.,\\ Mijnbouwstraat 120 2628 RX Delft
The Netherlands}

\maketitle

\begin{abstract}
This paper describes Team Delft's robot, which won the Amazon Picking
Challenge 2016, including both the Picking and the Stowing competitions. The
goal of the challenge is to automate pick and place operations in unstructured
environments, specifically the shelves in an Amazon warehouse. Team Delft's
robot is based on an industrial robot arm, 3D cameras and a customized gripper.
The robot's software uses ROS to integrate off-the-shelf components and
modules developed specifically for the competition, implementing Deep Learning
and other AI techniques for object recognition and pose estimation, grasp
planning and motion planning.
This paper describes the main components in the system, and discusses its
performance and results at the Amazon Picking Challenge 2016 finals.
\keywords{robotic system, warehouse automation, motion planning, grasping, deep
learning.}
\end{abstract}

\input{introduction}

\input{concept}

\input{software}

\input{vision}

\input{grasping}

\input{motion}

\input{results}

\input{conclusions}

\section{Acknowledgements}
All authors gratefully acknowledge the financial
support by the European Union's Seventh Framework Programme
project Factory-in-a-day (FP7-609206) We would like to thank 
RoboValley\footnote{\url{http://www.robovalley.com}}, the ROS-Industrial
consortium, our sponsors Yaskawa, IDS, Phaer, Ikbenstil and Induvac, the people at the Delft
Center for Systems and Control and TU Delft
Logistics for their support, also Lacquey B.V. for helping us handle our heavy rail, and
finally special thanks to Gijs vd. Hoorn for his help during the development of
the robotic system.

\bibliographystyle{splncs_srt}
\bibliography{apc}

\end{document}

%% file: introduction.tex
\section{Introduction}
The Amazon Picking Challenge (APC) was launched by Amazon Robotics in
2015 \cite{Correll-2016} to promote research into robotic manipulation for
picking and stocking of products.
These tasks are representative of the current challenges that warehouse
automation faces nowadays. The unstructured environment and the diversity of
products require new robotic solutions. Smart mechanical designs and advanced
artificial intelligence techniques need to be combined to address the
challenges in object recognition, grasping, dexterous manipulation or motion
planning. 

Amazon chose 16 teams from all over the world to participate in the finals at
RoboCup 2016. Team Delft won both the picking and
the stowing challenges.
Section 2 discusses Team Delft's approach, explaining its design principles and
the robot hardware. Section 3 details the robot control and all the components
integrated for object detection, grasp and motion planning. Finally sections 4
and 5 discuss the competition results and the lessons learned. The purpose of
this paper is to provide a comprehensive analysis of the complete development
of an advanced robotic system that has to perform in real-world
circumstances.

\section{The Amazon Picking Challenge 2016}

The Amazon Picking Challenge 2016 included two competitions: in the Picking
Task 12 items from the competition product set had to be picked from
an Amazon shelving unit and placed in a tote; in the Stowing Task it was
the other way around: 12 items were to be picked from the tote and stowed into
the shelf.
The maximum allotted time to fulfil each task was 15 minutes and the system had
to operate autonomously.
A file containing the task order was given to the system, which included the
initial contents of the shelf's bins and the tote, and it had to produce a
resulting file indicating the location of all the products.

The set of 39 items used in the challenge were representative of those in an
Amazon warehouse. Books, cubic boxes, clothing, soft objects, and
irregularly shaped objects represented realistic challenges such as reflective
packaging, different sizes or deformable shapes.
The items could be placed in any orientation inside the bins, sometimes 
cluttering them, and the target product could be partially occluded by
others.

Teams had to place their robots in a 2m x 2m workcell, no
closer than 10cm from the shelf. The workspace also posed important challenges
to perception and manipulation.
The shelf was a metal and cardboard structure divided into a matrix of 3 by 4
bins. The bins were narrow but deep, which limited the manoeuvrability inside
and required a long reach. Additionally, the shelf construction resulted in
significant deviations in reality from its ideal geometric model.

\begin{figure}
 \begin{center}
 \includegraphics[width=0.5\textwidth]{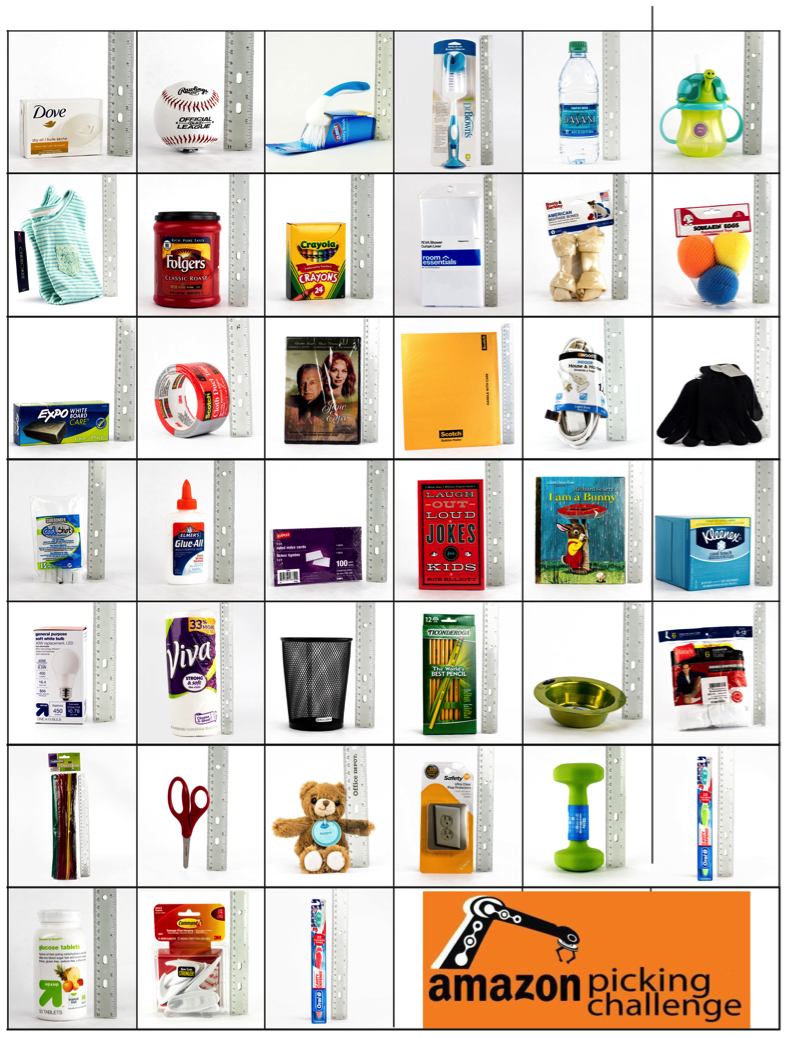}
 \end{center}
  \caption{The 39 items in the Amazon Picking Challenge 2016}
   \label{Fig:products}
\end{figure}

The performance of the robots during the picking and the stowing tasks was
evaluated by giving points for correctly placed items and subtracting penalty points for
dropping, damaging or misplacing items. A correct operation could receive 10, 15
or 20 points depending on the cluttering of the bin. Additional bonus points
were given for specially difficult objects, for maximum scoring of 185 points
in the Picking Task and 246 points in the Stowing Task.

%% file: concept.tex
\section{Team Delft's Robot}

Team Delft was a joint effort of the Robotics Institute of the Delft University
of Technology \cite{TUD-robotics} and the robot integrator company Delft
Robotics B.V. \cite{DelftRobotics} to participate in the APC 2016. 
Amongst TUD Robotics Institute research lines is the development of flexible
robots capable of automatizing small-scale productions, simplifying their
installation and reconfiguration, e.g. through automatic calibration or online
self-generated motions.
Delft Robotics is a novel systems integrator making these new robotic
technologies available to all kind of manufacturing companies. Both parties are
closely collaborating within the Factory-in-a-day EU project \cite{fiad} to
reduce installation time and cost of robot automation.
Team Delft's goal was to demonstrate and validate this approach in such a
challenging industrial benchmark as the APC. The team did not adapt and tune an
extant pick-an-place solution to participate in APC, but developed the best solution possible with
extant industrial hardware and as many off-the-shelf software components as
possible. For that the robot control was based on the ROS framework
\cite{Quigley-2009}.

The remaining of this section describes the main ideas behind Team Delft's
robotic solution.

\subsection{Robot Concept}
The team analysed the results of the previous edition of the Amazon Picking
Challenge in 2015 \cite{Correll-2016}, and decided that making the system
\emph{robust} and \emph{fast} was key to win. These characteristics allow
the system to perform several attempts to pick each target item, and move occluding objects
around if necessary.
We also learned that suction was the better performing grasp option, which we
confirmed in early tests.

\begin{figure}
 \begin{center}
 \includegraphics[width=0.7\textwidth]{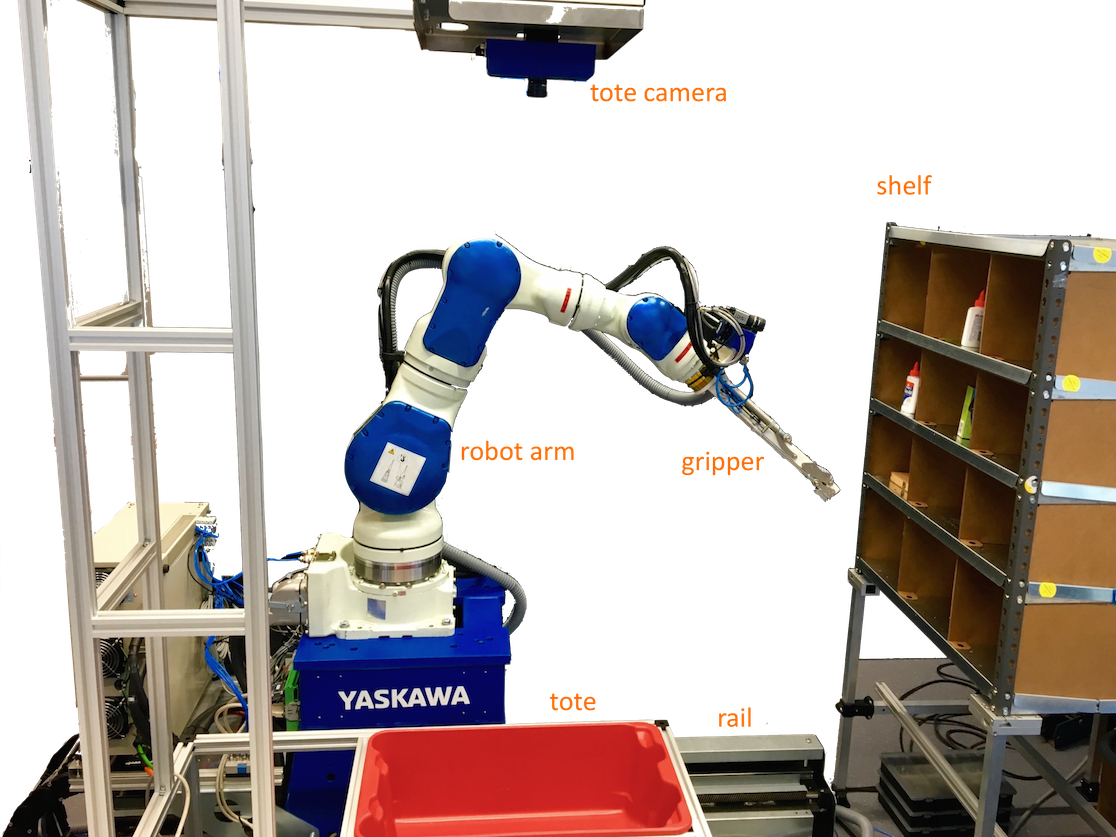}
 \end{center}
  \caption{Team Delft robot setup in the APC workcell.}
  \label{f:setup}
\end{figure}

The solution designed is based on an industrial robot arm, a custom made gripper
and 3D cameras, as shown in figure \ref{f:setup}. For the robot arm we chose
a 7 degrees of freedom SIA20F Motoman  mounted on an horizontal rail perpendicular to the shelf. The
resulting 8 degrees of freedom allowed the system to reach all the bins with
enough manoeuvrability to pick the target objects.
 
\begin{figure}
 \begin{center}
 \includegraphics[width=0.6\textwidth]{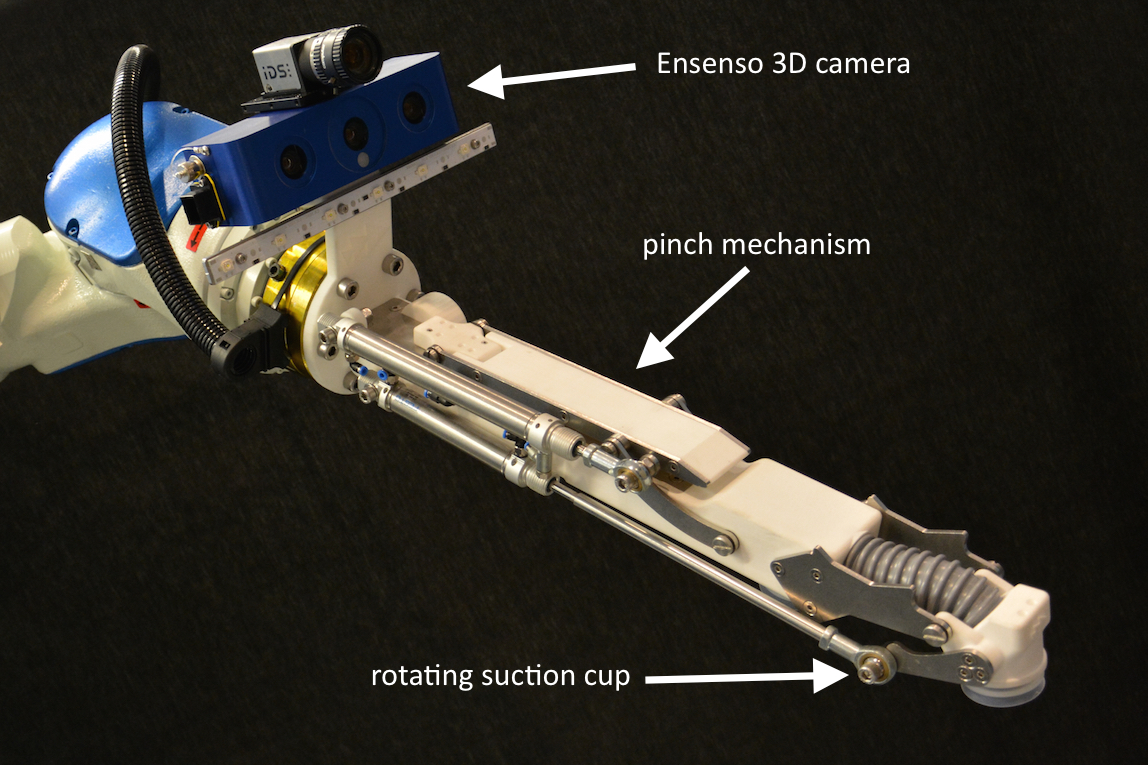}
 \end{center}
  \caption{Team Delft gripper.}
   \label{f:gripper}
\end{figure}

We customized our own gripper to handle all the products in
the competition (see figure \ref{f:gripper}). It has a lean footprint to
manoeuvre inside the bins, and a 40cm length to reach objects at the back.
It includes a high flow suction cup at the end, with a 90 degrees rotation
allowing two orientations, and a pinch mechanism for the products difficult to suck. Both
the suction cup rotation and the pinch mechanism are pneumatically actuated.
A vacuum sensor provides boolean feedback whether the suction cup holds
anything.
For object detection a 3D camera is mounted in the gripper to scan the bins,
while another one is fixed on a pole above the tote.

The tote is placed on a frame attached to the
robot rail. The compressor and the vacuum pump required to actuate the gripper
are mounted on another frame that attached to the rail base, so the whole set up
could be easily moved in three big blocks. Robust and easy transportation and installation were important
requirements.

\subsection{Control Pipeline}\label{ss:control}

The system control is based on the sense-plan-act paradigm and path planning
for robot motion.
This allows for potentially optimal motions, at the cost of more precise sensing
information.
First, the task is decomposed into a set of pick and place operations on the
target items.
Then, for each operation in the Picking task the sense-plan-act cycle proceeds
as follows\footnote{A video demonstrating the pipeline can be found here:
\url{https://www.youtube.com/watch?v=PKgFy6VUC-k}}. First in the \emph{sense}
step the robot moves to take an image of the bin containing the first target
item to locate it and get the obstacles information.
Then, during the \emph{plan} step a grasping
strategy and candidate pose for the gripper to grab it are computed, and a
motion plan is generated to approach, grasp and retreat from the bin with the
item. Following, in the \emph{act} step the gripper is configured for the
selected strategy and the complete motion is executed, including gripper activation to suck or
pinch-grasp the item.
The vacuum seal in the suction cup is checked to confirm a successful pick.
If so, the robot moves to deposit the item in the tote, using simple drop-off motions.
This cycle is repeated till all target items are picked. For the Stowing task
the loop operates similarly until all items in the tote are stowed in the shelf.

%% file: software.tex
\section{Robot Software}

Team Delft was fully committed to the ROS-Industrial initiative \cite{ROS-I}
that aims to create industry-ready, advanced software components to extend the
capabilities of factory robots. The robot software is thus based on the ROS
framework \cite{Quigley-2009}. We found that the flexibility, modularity and
tools provided by ROS allowed us to address the requirements for autonomy and
high and reliable performance in the competition, and facilitated development.

The ROS component-based approach allowed for the integration of the different
components for task management, object detection, pose estimation, grasping and
motion planning into a robust architecture. Following we describe them.

\subsection{Task Management}
On top of the architecture sits the task manager, responsible for decomposing
the Pick and the Stow tasks into a plan of pick and place operations, and
manages the state of fulfilment of the whole task. It encodes the competition
rules to maximize the scoring, by planning first those operations that scored
more points, and keeps track of the location of all the items.

A central coordinator module coordinates the execution of each pick and place
operation following the sequential flow presented in section \ref{ss:control}.
It was implemented as a ROS SMACH \cite{Smach} state machine.

The system can handle some failures applying fallback mechanisms to 
continue operation. For example, if the robot cannot find the target item, or
estimate its pose, it tries different camera viewpoints, then if the problem
persists it postpones that target and moves to the next operation.
The system can detect if a suction grasp failed by checking the vacuum
sealing after execution of the complete grasp and retreat action. If there is no
seal the robot assumes the item dropped inside the bin and retries the pick
later.
If the seal is broken during the placement in the tote, the item is assumed to
have dropped in the tote.

%% file: vision.tex
\subsection{Object Recognition and Pose Estimation}

The robot's pipeline involved detecting the target item within the bin or the
tote and obtaining a grasp candidate using an estimation of its pose or its
centroid, in the case of deformable items. Difficulties included narrow view
angles and poor lighting conditions and reflections inside the shelf.

Firstly, the system acquires the 3D\footnote{The 3D data format used was
PointCloud.} and RGB image with an Ensenso N35 camera.
For that, in the Picking task the robot moves the gripper to a pre-defined
location in front of the desired bin. In the Stowing task the image is taken by
the camera fixed over tote. Then object detection consists of two main
steps:

\subsubsection{Object Recognition.}
\begin{figure}
\begin{center}
 \includegraphics[width=0.4\textwidth]{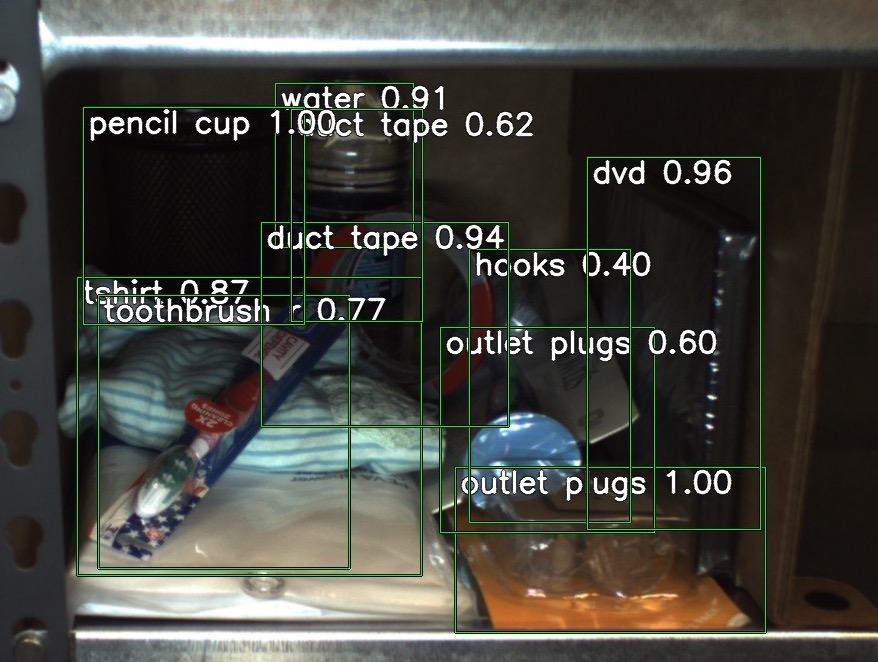}
 \includegraphics[width=0.48\textwidth]{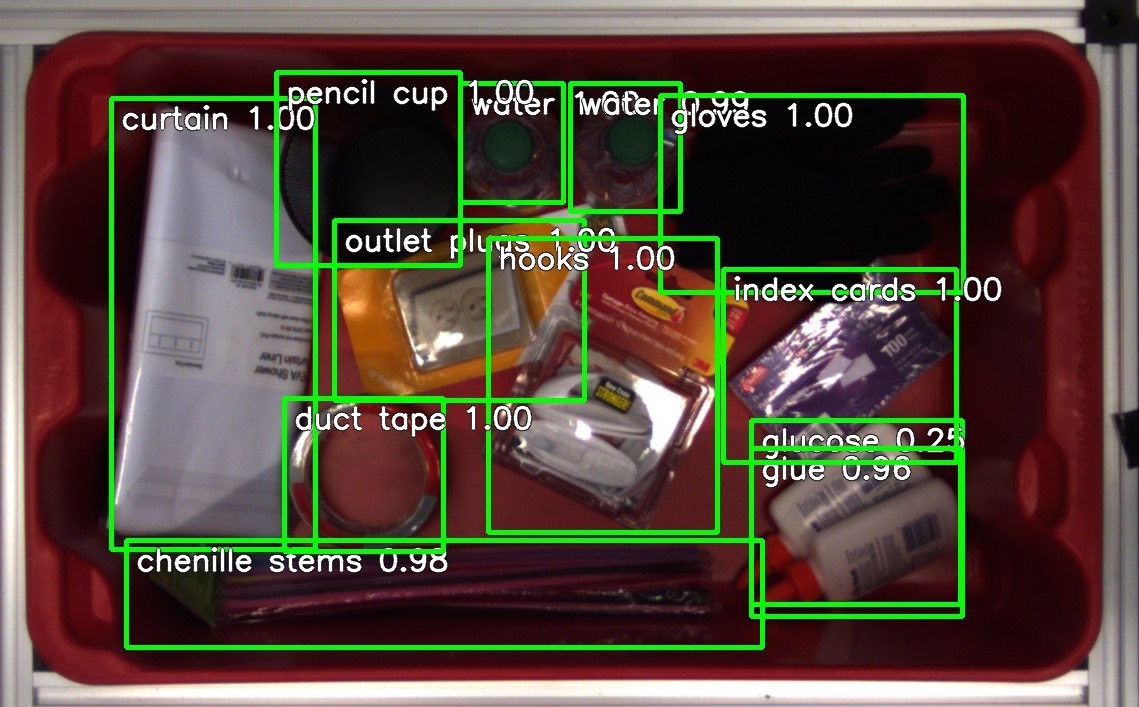}
 \end{center}
  \caption{Example result of the object detection module based on Faster R-CNN
  for bin and tote images.  The estimated bounding boxes are depicted in green
  and labelled with the identification proposal and the confidence.}
   \label{f:object_detection}
\end{figure}

First, a deep neural network based on Faster
R-CNN \cite{Ren-2015} classifies the objects in the RGB image and extracts their
bounding boxes. A pre-trained neural network was further trained to create the two models used for object recognition in both the picking and the stowing tasks.
A dataset of about 20K images of the products in different orientations
and with random backgrounds was created to train a ``base'' model. Then this
model was trained with around 500 labelled images of real occurrences of the
products in the shelf and in the tote to generate the final recognition models.
The result was an almost flawless detection within 150ms of all the products
present in any bin or tote image, as shown in figure \ref{f:object_detection}.

\subsubsection{Pose Estimation.} Pose estimation of non-deformable products
was done using Super 4PCS \cite{Mellado-2014} to match the filtered PointCloud of the target item with a
CAD model of the object. The 3D information of the bin or the tote is also used later during
motion planning for collision detection. Reflections due to packaging
and the difficult lighting conditions inside the bin resulted in scarce and
noisy 3D data for some products. This proved a big difficulty for the pose estimation
method. We included heuristics to correct estimations, e.g. objects
cannot be floating on the bin, and also the mentioned fall-back mechanism to
take additional images.

%% file: grasping.tex
\subsection{Grasping and Manipulation}
\begin{figure}

\begin{center}
 \includegraphics[width=0.4\textwidth]{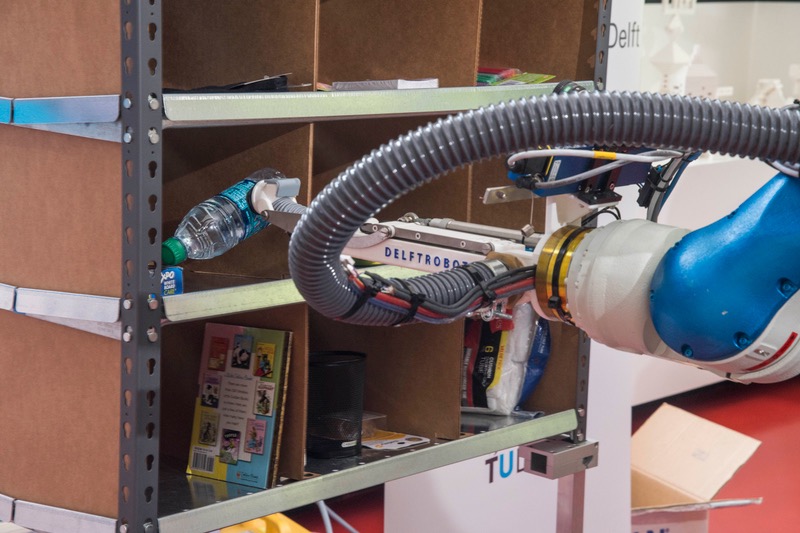}
 \includegraphics[width=0.4\textwidth]{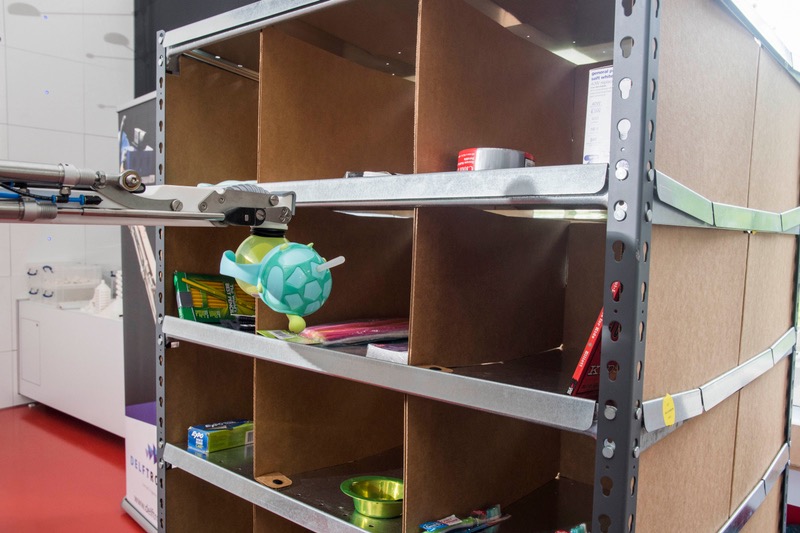}
 
  \includegraphics[width=0.8\textwidth]{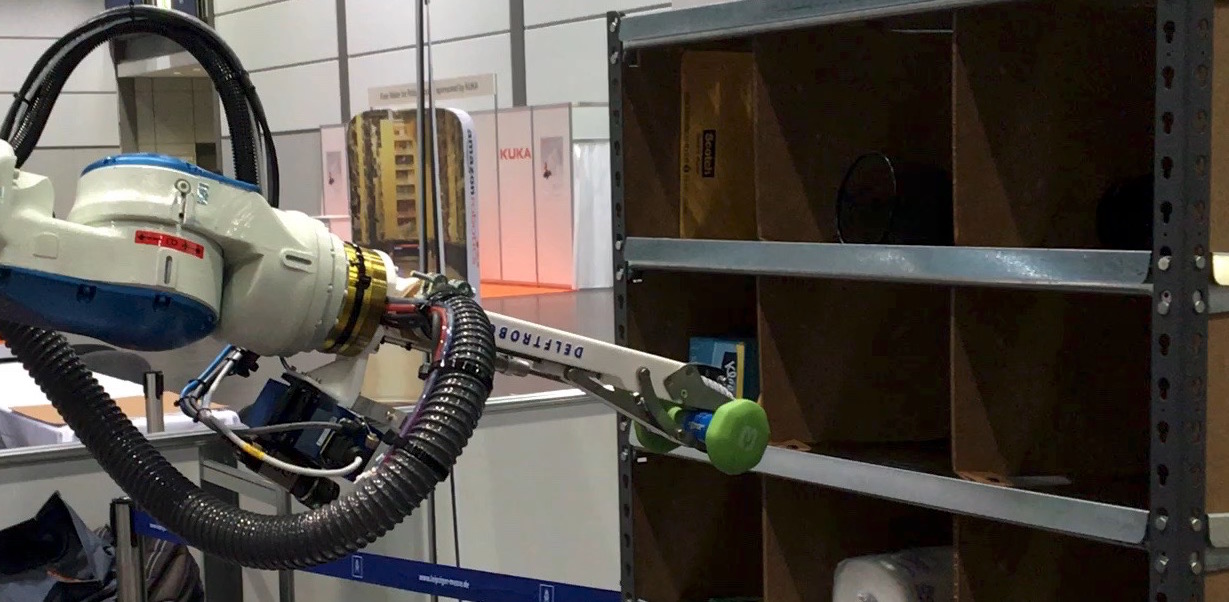}
 \end{center}
  \caption{The different gripper configurations for the grasping strategies. In
  the top images, the two configurations for suction rotating the cup. In
  the image below, the robot is picking the dumbbell using the pinch
  configuration.}
   \label{fig:grasping}
\end{figure}

The grasp and manipulation solution is customised to our gripper and our
path planning approach.
The gripper has three basic modes or configurations
(see figure \ref{fig:grasping}): \emph{front suction}, \emph{side-top suction},
and \emph{pinch}, each one corresponding to a grasping strategy more suitable
different products also depending on the situation.

In the \emph{plan} step the best strategy and associated
grasp candidate --i.e. a 6D pose to position the gripper-- to grasp the target
item are chosen, and then the system computes a manipulation plan to move the
gripper to the candidate pose, activate it to pick the item, and move
out of the bin (or the tote) holding it.

\subsubsection{Grasp planning.}
For non-deformable items pose estimation of the object and
offline geometric constraints and user-defined heuristics are used to synthesize
 grasp candidates.
Basically a set of grasp candidates is generated over the surface of the 3D
model of the item based on primitive shapes (cylinders, spheres, cones,
planes, etc.). These candidates are pruned online using geometry constraints
due to the actual item's estimated pose and gripper limitations, e.g. candidates at the
back or the bottom\footnote{These are relative locations assuming we are
looking at the object from the tip of the gripper.} of the item are not reachable.
Additional heuristics were defined experimentally to prune those grasp
candidates specific to each product that proved not suitable. These heuristics
were implemented so that new ones could easily be coded for a set of products by
including them in the primitive shapes that accounted for different products,
while still being able to define ad-hoc constraints that only applied to
specific products. Finally, the synthesized candidates are scored based on
geometry and dynamic considerations, e.g. poses closer to the centre or mass
would tend to provide more stable grasps.

For deformable items the system exploits the power of the suction
cup, which is capable of grasping and holding all
deformable products in the competition. Instead of computing grasp candidates
from the 3D pose estimation of the object, the normals of the segmented object
PointCloud are directly used as grasp candidates. The candidates are also
scored simply based on the distance to the PointCloud centroid, the closer the
better.

\subsubsection{Manipulation.}
Actually grasp planning produces not one but a set of grasp candidates ranked by
our scoring criteria. However, the robot might not be able to reach some of
these poses with the gripper, due to its kinematic limits or the obstacles, such as
the bin or the tote walls, or other items close to the target one. Even if
reachable, the robot, with the item attached to
the gripper, also needs a retreat trajectory  free of obstacles. The
first grasp candidate for which a collision-free pick and retreat complete trajectory can be computed is then
selected. This will be detailed in section \ref{sss:plan}.

%% file: motion.tex
\subsection{Motion Planning}
For the robot motion strategy we divided the problem considering that the
workspace is static (apart from motions due to the robot), and known outside of
the shelf. Any online motion planning was
required only inside the bins or the tote.

\subsubsection{Offline Motions.}
Collision-free trajectories between all
relevant locations to approach the bins and the tote and capture the images
were computed offline.
The trajectories were generated in joint space with the RRT-Connect randomized path
planner via MoveIt! \cite{MoveIt} and using a URDF\footnote{Unified Robot
Description Format \url{http://wiki.ros.org/urdf}} model of the workcell
including the shelf, the robot on the rail, the gripper, and all the attached frames and equipment.

\subsubsection{Online Cartesian Path Planning.}\label{sss:plan}
To simplify the manipulation problem inside the bins, only collision-free picks
were to be attempted. We defined a cartesian approach based on the MoveIt! pick
and place pipeline that took the target grasp candidate and computed a
combination of linear segments to \emph{approach}, \emph{contact} grasp the target object, \emph{lift} it after grasping and
\emph{retreating} with it. The TRAC-IK library \cite{Beeson-2015} is used for
inverse kinematics, configured to enforce minimal configuration changes, and then
collision checking is done with MoveIt! using the PointCloud
information from the camera.

\subsubsection{Robot Motion.}
This way, for the Picking task, offline motions were used in the \emph{sense}
phase to acquire the image of the bin containing the target object and to
position the gripper ready to enter the bin.
Then, during the \emph{plan} phase the approach, contact, lift and retreat
segments were generated online, and a drop-off location chosen and  an
associated offline trajectory retrieved. Finally, a complete motion plan to pick
and place the target item is generated by stitching the cartesian segments and
the offline drop-off trajectory. This includes time parametrization and the I/O
commands required to configure and activate the gripper for grasping, resulting
in a complete trajectory that is executed by the robot in the \emph{act} phase.
The MotoROS driver was used and enhanced by Team Delft\footnote{This
contribution, as well as other ROS components developed for APC will be open
to the community.} to execute the desired trajectories controlling the complete
kinematic chain of the robot and the rail, and also the gripper using the robot controller I/O.

%% file: results.tex
\section{Competition Results}

Team Delft's robot was the champion of the  challenge winning both competitions,
with an outstanding performance in the Stowing Task\footnote{Video recordings of Team Delft's competition runs
can be found here \url{https://youtu.be/3KlzVWxomqs} (picking) and here
\url{https://youtu.be/AHUuDVdiMfg} (stowing)}.
Table 1 shows the final scores for
the Amazon Picking Challenge 2016 Pick and Stow competitions. 
The overall results of the teams improved considerably over the previous APC
edition: average scoring for the top 10 teams increased 38\% for the Picking
Task, specially considering the increased difficulty in this edition, with more
cluttered bins.
It is also interesting to mention that all the best robots but Team Delft's
placed the tote below the shelf, and initially moved a board to act as a ramp so
that any items dropping will fall down to the tote. This trick improved scoring. We considered this mechanical
solution early at the concept brainstorming, but finally discarded it because
due to the rail there was no free space for a clean and robust design. We did
not want to include any provisional duck-tape solution.
However, Team Delft's robust and fast concept outperformed the rest achieving
more successful pick and place operations, which was the aim of the
competition.

In the Stowing Task Team Delft's robot successfully stowed 11 items of the 12,
dropping the remaining one while manipulating one of the other products. The
system only had to retry one of the picks from the tote, to finish the task in
a total time of 7min 10secs.
\begin{table}
\caption{Amazon Picking Challenge 2016 scores of the best four robots.}
\begin{center}
\begin{tabular}{cl}
\hline
& \textbf{Stowing scores} \\
\hline
\textbf{214}		&	Team Delft	  \\
\textbf{186}	 	&	NimbRo Picking	 	\\
\textbf{164}		&	MIT					\\
\textbf{161}		&   PFN					\\
\hline
\end{tabular}
\quad
\begin{tabular}{cl}
\hline
& \textbf{Picking scores} \\
\hline
\textbf{105}		&	Team Delft (0:30  first pick) \\
\textbf{105}	 	&	PFN	(1:07 first pick) \\
\textbf{97}		&	NimbRo Picking			\\
\textbf{67}		&   MIT				\\
\hline
\end{tabular}
\end{center}
\label{t:results}
\end{table}

The picking task proved much harder than the stowing. The robot picked
successfully 9 out of 12 items, the first one in only 30 secs. The robot
dropped one of the targets and was not able to pick the remaining two. The
system dropped a non-target item during manipulation. The system also
successfully moved 5 items between the shelf's bins to clear occlusions for required picks. Two of those move operations allowed it two
successfully pick two target products. Team Delft called the end of the run
after 14min and 45secs.

\subsection{Analysis}
Team Delft's robot reliable and performing capabilities were the key to its
success. Its gripper could grasp all 39 items in
the competition, including the dumbbell and the pencil cup using the pinch
grasp, in any orientation and bin location. However, the grasp on heavy and big items was not
completely reliable. The dumbbell proved specially difficult, since the grasp
approach needed extreme accuracy to succeed.

The object recognition module had an specially outstanding performance robust to
varying light conditions. However, pose estimation was
strongly affected by reflections, which produced scarce PointCloud data.

Most difficulties for
our system were encountered when trying to find a collision-free motion plan to
approach the target object. This rejected many targets that were retried
later. In the next attempt, removing occluding items was done, but sometimes the
cluttering of the bin caused a stall situation in which items were preventing
each other from being picked.

Overall, the Picking Task proved far more difficult than the Stowing Task, with
many teams scoring half the points. This is because picking from the shelf
required more manipulation, with items occluding each other. The Stowing task was basically a
standard bin picking problem: all items in the tote were to be picked, and
gravity helps having some easily accessible at the top.
Also, the stowing in the shelf could be done with pre-computed motions to shove
the target item in the bin, blindly pushing back any previous content.

%% file: conclusions.tex
\subsection{Lessons Learned}
Considering the results described in the previous section and the complete
experience developing the robot for the Amazon Picking Challenge, we reached
several conclusions about our concept design premises and how to improve it.

The most important idea is that manipulation requires contact with the
environment. Team Delft's pure planning approach to grasping and manipulation
treated contact as collisions to avoid, and simply by-passed this constraint for
the target object. This caused a lot of rejected plans to grasp items from
cluttered bins, some of them becoming actually unrealisable. Force-feedback and
compliance in the gripper seem unavoidable to achieve a reliable solution.
 Also, creating a single gripper capable of
handling such a variety of products proved difficult. None of the teams managed to pick the dumbbell, for example. Having
different grippers and switching between them on the fly seems a more efficient
and robust solution.

On the perception side, Deep Learning neural networks proved an excellent
solution for object recognition, but they also are a really promising solution
to pose estimation and even grasp planning, as the results of other teams
suggest. 

Notwithstanding the discussed improvements, Team Delft's concept based on speed
and reliability proved successful.
The ready-for-industry approach we took, with installation and setup
procedures, and professional team coordination during the competition, 
allowed to keep robustly improving the robot's performance till reaching close
to its top limit right at the competition.

\section{Concluding Remarks}
This paper provides a comprehensive overview of Team Delft's robot winner of the
Amazon Picking Challenge 2016. The key to Delft's robot success was a concept
aimed for robustness and speed, relying on an end-to-end engineering process
integrating well establish industry practices and cutting-edge AI and robotics
technologies.

There was a new Stowing Task in the 2016 edition of the challenge, to bin-pick
products from a tote and stow them in a shelf.
The overall high scores by many teams, and the excellent
performance of Team Delft's robot, suggest that the bin picking problem for
diverse, medium-size products can be addressed by current robotic technology.
Speed is still far from human performance (\~100 items an hour, compared to 400
items an hour in the case of a human), but considering that Team Delft's robot
could have been speed-up probably 50\% with faster motions and faster
processing, we are confident to predict that robot technology is getting there.
However, the Picking task results, proved that general manipulation, including
diverse objects and cluttered spaces, still remains an open problem for
robotics.